\pdfoutput=1
\documentclass[11pt]{article}

\usepackage[]{acl}

\usepackage{times}
\usepackage{latexsym}

\usepackage[T1]{fontenc}
\usepackage[utf8]{inputenc}

\usepackage{microtype}

\usepackage{inconsolata}
\usepackage{graphicx}
\usepackage{tabularx}
\usepackage{soul}

\usepackage{multirow}
\usepackage{amsmath,amssymb,amsfonts}
\usepackage{amsthm}
\usepackage{mathrsfs}
\usepackage[title]{appendix}%
\usepackage{xcolor}%
\usepackage{textcomp}%
\usepackage{manyfoot}%
\usepackage{booktabs}%
\usepackage{algorithm}%
\usepackage{algorithmicx}%
\usepackage{algpseudocode}%
\usepackage{listings}%
\usepackage{float}
\usepackage{makecell}
\usepackage{ulem}
\usepackage{bbding}
\usepackage{xstring}
\usepackage{hyperref}

\title{Triad: A Framework Leveraging a Multi-Role LLM-based Agent to Solve Knowledge Base Question Answering}

\makeatletter
\def\@fnsymbol#1{\ensuremath{\ifcase#1\or \or \ddagger\or
\mathsection\or \mathparagraph\or \|\or **\or \dagger\dagger
\or \ddagger\ddagger \else\@ctrerr\fi}}
\makeatother

\author{Chang Zong$^1$, Yuchen Yan$^1$, Weiming Lu${^1}{^\dagger}$ \thanks{$^\dagger$Corresponding authors.}, Jian Shao$^1$ \\ \textbf{Yongfeng Huang$^2$, Heng Chang$^3$, Yueting Zhuang${^1}{^\dagger}$ \footnotemark[1]} \\ $^1$College of Computer Science and Technology, Zhejiang University \\ $^2$The Chinese University of Hong Kong \\ $^3$Tsinghua University \\
\texttt{\{zongchang, luwm, yzhuang\}@zju.edu.cn}
}

\begin{document}
\maketitle
\begin{abstract}
Recent progress with LLM-based agents has shown promising results across various tasks. However, their use in answering questions from knowledge bases remains largely unexplored. Implementing a KBQA system using traditional methods is challenging due to the shortage of task-specific training data and the complexity of creating task-focused model structures. In this paper, we present \textbf{Triad}, a unified framework that utilizes an LLM-based agent with multiple roles for KBQA tasks. The agent is assigned three roles to tackle different KBQA subtasks: agent as a generalist for mastering various subtasks, as a decision maker for the selection of candidates, and as an advisor for answering questions with knowledge. Our KBQA framework is executed in four phases, involving the collaboration of the agent's multiple roles. We evaluated the performance of our framework using three benchmark datasets, and the results show that our framework outperforms state-of-the-art systems on the LC-QuAD and YAGO-QA benchmarks, yielding F1 scores of 11.8\% and 20.7\%, respectively.
\end{abstract}

\section{Introduction}
A question-answering system is designed to extract information by converting a natural language question into a structured query that can retrieve precise information from an existing knowledge base \cite{KGQAN}. The resolution of Knowledge Base Question Answering (KBQA) typically involves phases including question understanding, URI linking, and query execution. Traditional KBQA systems require the use of specialized models trained with domain datasets for question parsing and entity linking \cite{gAnswer,KGQAN,EDGQA}. Large language models (LLMs), however, have shown promising competencies in in-context learning using task-specific demonstrations \cite{ICLSurvey}. LLMs have recently been employed as agents in the execution of complex problems. A framework that employs LLM-augmented agents can generate actions or coordinate multiple agents, thus improving the capacity to handle complex situations \cite{BOLAA}. Despite the remarkable performance of LLMs in various tasks as evidenced in previous studies, a comprehensive qualitative and quantitative evaluation of KBQA frameworks empowered with an LLM-based agent remains insufficiently explored.

\begin{figure}[htbp]
    \begin{center}
    \includegraphics[width=.4\textwidth]{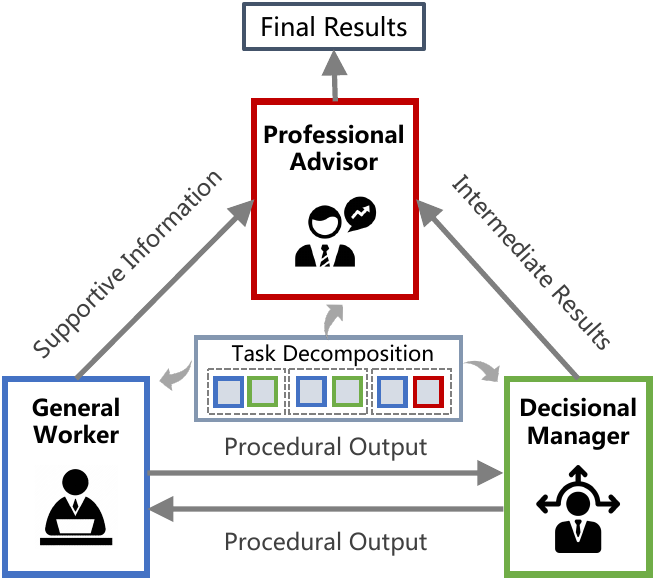} 
    \caption{A system with multiple roles who focus on sub-problems of each phase to solve a complex task.}
    \label{multirole}
    \end{center}
\end{figure}

Studies on KBQA with LLMs has attracted considerable attention. Some works focus primarily on highlighting the inability of LLMs to generate complete factoid results \cite{Empirical,GPTEval} or demonstrating their potential efficacy in future research \cite{ChatGPT-KBQA,CanCR}. Other works concentrates on generating answers by prompt learning and incorporating external knowledge bases \cite{PromptING,McL-KBQA}. Concurrently, LLMs can be deployed to address each phase within Text2SQL challenges\cite{Graphix-T5,BIRD} or theorem proof tasks\cite{LLMScience}. However, each phase of KBQA can be further decomposed into subtasks and completed through an agentic approach that provides feedback and cooperation. Additionally, decomposing the task reduces the complexity of cooperative working by allowing each role to concentrate on smaller sub-problems\cite{wang2020rode}. As illustrated in Figure \ref{multirole}, three roles in an organization work together to provide the final answer for the overall task. The above observations spur our exploration into the following question: \textbf{How does an LLM-based agent solve KBQA tasks by serving as multiple roles, and its performance is comparable to systems trained specifically?}

In this study, we introduce \textbf{Triad}, a unified framework that leverages an LLM-based agent with three roles to address KBQA tasks. Specifically, we implement the agent consisting of an LLM as the core, supplemented by various task-specific modules such as memory and executing functions. The agent is assigned three distinct roles: a generalist (G-Agent) adept at mastering numerous small tasks by the given examples, a decision maker (D-Agent) proficient at identifying options and selecting candidates, and an advisor (A-Agent) skilled at providing answers using internal and external knowledge. The cooperation of these agent roles composes a KBQA process containing four phases: question parsing, URI linking, query construction, and answer generation. We evaluate our framework on three benchmark datasets in various difficulties. The results show that our framework outperforms state-of-the-art systems, demonstrated by 11.8\% and 20.7\% F1 scores on the LC-QuAD and YAGO-QA benchmarks, respectively\footnote{Code and data are available at \url{https://github.com/ZJU-DCDLab/Triad}.}. 

The contributions of this study can be summarized as follows: 
\begin{itemize}
    \item We propose Triad, the first framework that leverages an LLM-based agent to solve KBQA tasks in all its four phases, without specialized training models.
    \item We implement an LLM-based agent with various task-specific modules that can act as three roles, including a generalist, a decision maker, and an advisor, to collaboratively solve KBQA via focusing on subtasks.
    \item We evaluate the performance of Triad. The results show a competitive ability compared to both state-of-the-art KBQA systems and pure LLM methods.
\end{itemize}

\section{Preliminaries}
\subsection{Phases of KBQA}
A typical KBQA system has a process that encompasses four phases:

\paragraph{Question parsing} involves converting natural language questions into a structured format that incorporates references to entities and relations.
\paragraph{URI linking} entails associating and replacing these entity and relation mentions with their corresponding URIs within a knowledge base. 
\paragraph{Query construction} involves creating executable queries in a standard format to extract answers from knowledge bases.
\paragraph{Answer generation} seeks to obtain the ultimate answers either by performing queries within knowledge bases or by directly querying an agent.

\subsection{Roles of LLM-based Agent}
Drawing an analogy to a software development scenario, where coders complete small development tasks, with the process and plan being decided by the manager, and ultimately the outcome inspected by the leader, we assign the following three roles to an LLM-based agent to solve the KBQA task:

\paragraph{Agent as a generalist} (G-Agent) is capable of mastering various small tasks by providing a few examples.
\paragraph{Agent as a decision-maker} (D-Agent) adepts at analyzing options and providing candidate results as procedural feedback.
\paragraph{Agent as an advisor} (A-Agent) is skilled in providing final answers with the aid of both external and its own knowledge. 

\subsection{Task Formulation}
A KBQA task refers to the process of solving a set of subtasks $S$. Each subtask $S_t \in S$ contributes to one phase of the whole process. An LLM-based agent $Agent_r$ with a role $r$ can be used to resolve a type of subtasks by its task-specific components, including a language model $LLM$, a memory $Mem_t$, a function $F_t$, a prompt $Pmt_t$ and a set of parameters $\theta_t$, using the set of role-related hyperparameters $\sigma_r$. The task can be formulated as follows:

\begin{equation}
\begin{aligned}
&f(KBQA) = \mathop{\bigoplus}\limits_{t=1}^T f(S_t) \\
f(S_t) =& Agent_r(LLM, Mem_t, F_t, Pmt_t, \theta_t, \sigma_r)
\label{eq:eq1}
\end{aligned}
\end{equation}, where $T$ is the total number of subtasks, $\bigoplus$ is the way to coordinate subtasks to solve the whole.

\section{Triad Framework}
\begin{figure*}[htbp]
    \begin{center}
    \includegraphics[width=1.02\textwidth]{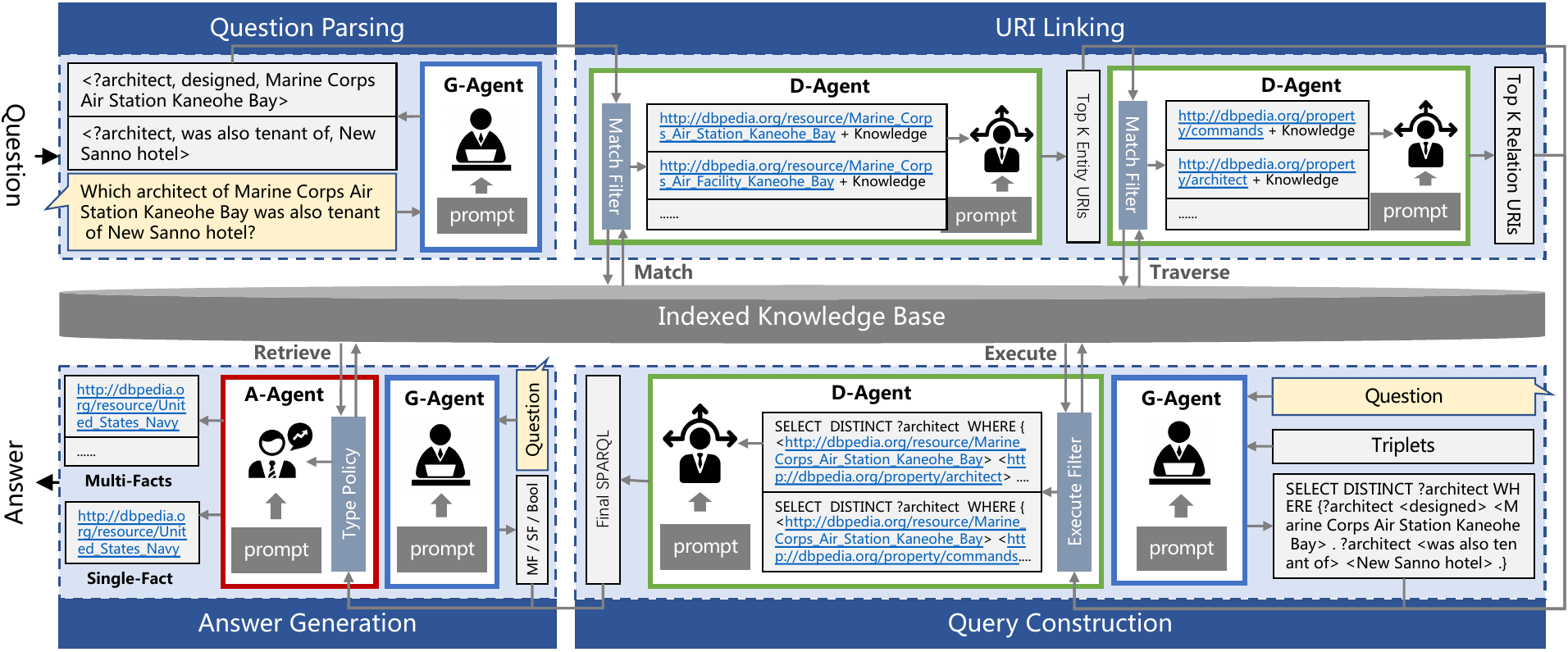} 
    \caption{Our Triad framework leverages an LLM-based agent with three different roles including a generalist, a decision-maker, and an advisor to cooperatively handle a series of subtasks in the four phases of a KBQA process.}
    \label{architecture}
    \end{center}
\end{figure*}

The overall architecture of \textbf{Triad} is shown in Figure \ref{architecture}. Each role of the LLM-based agent, along with its associated subtasks, is illustrated as follows.

\subsection{G-Agent as a Generalized Solver}
A generalized agent (G-Agent) proficiently manages numerous tasks by leveraging learning from limited examples through an LLM. In our framework, a G-Agent can perform question parsing, query template generation, or answer type classification as actions solely utilizing an LLM. These three subtasks are illustrated as follows:

\paragraph{Triplet mention extraction:} The process of extracting triplet mentions in question parsing involves the conversion of a naturally phrased question, denoted as $Q$, into formatted triplets of entities and relations. This subtask is executed employing an LLM, which is guided by a prompt with a set of prerequisites and a selection of examples. This subtask can be represented as follows:
\begin{equation}
\begin{aligned}
f(S_{tri}) =& Agent_g(LLM, Pmt_{tri}, Q, \mathcal{N}) \\ 
Pmt_{tri} =& \left[Ins_{tri}, Shot_{tri}, CoT_{tri}\right]
\label{eq:eq2}
\end{aligned}
\end{equation}, where $Agent_g$ is the agent as a generalist to perform the triplet extraction subtask with $\mathcal{N}$ examples. $Pmt_{tri}$ is the prompt to guide $LLM$ to generate triplets from the question $Q$, which consists of instruction $Ins_{tri}$, examples $Shot_{tri}$, and chain-of-thought prompt $CoT_{tri}$ \cite{CoT}.

\paragraph{SPARQL template generation:} The generation of SPARQL templates in query construction involves the use of an LLM to create a SPARQL template that articulates the question using standard SPARQL syntax, replacing URI identifiers with entity and relation variables. To derive precise and comprehensive answers from the knowledge base using SPARQL queries, there are two potential strategies. One approach involves the direct generation of an executable SPARQL using an LLM, though this method may significantly increase LLM call times and error rates when numerous candidate queries are in play. Alternatively, a SPARQL template can initially be generated with entity and relation variables, which are subsequently replaced with linked URIs. For the sake of stability and efficiency, we opt for the second strategy. This subtask can be denoted as:
\begin{equation}
\begin{aligned}
f(S_{qt}) =& Agent_g(LLM, Pmt_{qt}, \theta_{qt}, \mathcal{N}), \\
Pmt_{qt} =& \left[Ins_{qt}, Shot_{qt}, CoT_{qt}\right], \\
\theta_{qt} =& \left[Q, f(S_{tri})\right]
\label{eq:eq3}
\end{aligned}
\end{equation}, where $Agent_{g}$ is the agent as generalist to perform SPARQL template generation with $\mathcal{N}$ examples, $f(S_{tri})$ is the triplets derived from the previous subtask, $Pmt_{qt}$ is the prompt for $LLM$ to generate SPARQL template.

\paragraph{Answer type classification:} In the phase of answer generation, the answer type classification subtask refers to the process of assigning a specific category to a response according to the question. This process serves as a guiding mechanism for the framework to generate comprehensive and accurate answers. This classification subtask is denoted as:
\begin{equation}
\begin{aligned}
f(S_{cls}) =& Agent_g(LLM, Pmt_{cls}, Q, \mathcal{N}), \\ 
Pmt_{cls} =& \left[Ins_{cls}, Shot_{cls}, CoT_{cls}\right]
\label{eq:eq4}
\end{aligned}
\end{equation}, where $Agent_g$ is the agent as a generalist to perform type classification subtask with $\mathcal{N}$ examples, $Pmt_{cls}$ is the prompt for $LLM$.

\subsection{D-Agent as a Decision-Maker}
An agent as a decision maker (D-Agent) is capable of making candidate selections step by step through filtering and choosing from given options, harnessing the capabilities of an LLM and KB as memory. The D-Agent can effectively handle three subtasks, which are delineated as follows:

\paragraph{Candidate entity selection:} The selection of candidate entities in URI linking is pivotal to the ultimate efficacy of KBQA. Prior research has focused primarily on developing a semantic similarity model to address this linking challenge. However, the linking task requires numerous iterations of searching within the knowledge base, which poses a compatibility issue for LLM-oriented methods. In our framework, an agent as a decision maker is utilized initially to filter all potential entity URIs from the knowledge base, subsequently deploying an LLM to select candidate URIs from a pool of potential identifiers. For each entity, our aim is to find the $\mathcal{K}$ most possible entity URIs which can be used to traverse over the KB to get the final answer. The entity selection subtask can be denoted as:

\begin{equation}
\begin{aligned}
f(S_{es}) =& Agent_d(LLM, Mem_{es}, F_{es}, \\
&Pmt_{es}, \theta_{es}, \mathcal{K}), \\ 
Mem_{es} =& \left[KB, List_{es}\right], \theta_{es} = \left[Q, f(S_{tri})\right]
\label{eq:eq5}
\end{aligned}
\end{equation}, where $Agent_d$ is the agent as a decision maker to perform the entity selection subtask with question $Q$, extracted triplets $f(S_{tri})$ and memory $Mem_{es}$, $Mem_{es}$ is composed of a knowledge base $KB$ and a list of entity URIs $List_{es}$ filtered from $KB$ using a text similarity matching function $F_{es}$, $Pmt_{es}$ is the prompt for LLM to perform the subtask, $\mathcal{K}$ is the hyperparameter of $Agent_{d}$, indicating the number of candidates selected by $LLM$.

\paragraph{Candidate relation selection:} The task of candidate relation selection in URI linking presents considerable challenges due to the discrepancies between word forms and meanings.  Nevertheless, the existence of reasoning paths in the knowledge base can be utilized to allow for a significant reduction of the search space in relation linking. In our framework, an agent as a decision maker endeavors to sieve through all potential relation URIs by navigating the knowledge base with candidate entity URIs generated from the previous subtask. Subsequently, an LLM is used to select the top $\mathcal{K}$ most probable relation URIs for output. The relation selection subtask can be denoted as:
\begin{equation}
\begin{aligned}
f(S_{rs}) =& Agent_d(LLM, Mem_{rs}, F_{rs}, \\
&Pmt_{rs}, \theta_{rs}, \mathcal{K}), \\ 
Mem_{rs} =& \left[KB, List_{rs}\right], \theta_{rs}=\left[Q, f(S_{es})\right]
\label{eq:eq6}
\end{aligned}
\end{equation}, where memory $Mem_{rs}$ is composed of the knowledge base $KB$ and a list of possible relation URIs $List_{rs}$ filtered from $KB$ using a one-order traversing function $F_{rs}$. $Pmt_{rs}$ is the prompt for LLM to perform relation selection. $\mathcal{K}$ is the number of relation URIs selected by LLM.

\paragraph{Candidate SPARQL selection:} The subtask of candidate SPARQL selection in query construction involves determining the appropriate SPARQL queries to obtain the final answers. Given a SPARQL template generated by the G-Agent, along with multiple candidate URIs selected from the D-Agent in previous subtasks, our D-Agent is targeted to identify the most plausible query. To further reduce the difficulty of selection, an executor function is applied to eliminate queries that cannot retrieve any results from the knowledge base. In conclusion, our aim in this subtask is to use D-Agent to construct executable SPARQLs and find the most possible one given a query candidate list with supported information. The SPARQL selection subtask can be denoted as:
\begin{equation}
\begin{aligned}
f(S_{qs}) =& Agent_d(LLM, Mem_{qs}, F_{qs}, \\
&Pmt_{qs}, \theta_{qs}, \mathcal{K}), \\ 
Mem_{qs} =& \left[KB, List_{qs}\right], \\
\theta_{qs}=&\left[Q, f(S_{es}), f(S_{rs}), f(S_{qt})\right]
\label{eq:eq7}
\end{aligned}
\end{equation}, where memory $Mem_{qs}$ is composed of a knowledge base $KB$ and a list of possible SPARQLs $List_{qs}$ constructed with SPARQL template $f(S_{qt})$, entity URIs $f(S_{es})$, and relation URIs $f(S_{rs})$ by the function $F_{qs}$, $Pmt_{qs}$ is the prompt for LLM to perform query selection, $\mathcal{K}=1$ is the number of queries selected by LLM.

\subsection{A-Agent as a Comprehensive Advisor} 
An advisory agent (A-Agent) is capable of processing a question and a corresponding type of answer as input. Its response is generated by either extracting information from an external knowledge base or by utilizing its internal knowledge to provide a direct answer. This comprehensive answering subtask can be described as follows:

\paragraph{Comprehensive answering:} The objective of comprehensive answering in the answer generation phase is to derive a definitive response based on an incoming question. Previous work \cite{ChatGPT-KBQA} has demonstrated that LLMs are more proficient in delivering single-fact responses and making Boolean judgments. Given this understanding, we implement an advisory agent that incorporates a simple policy to facilitate a comprehensive answering approach. Specifically, if a question yields a final SPARQL generated from the preceding steps, A-Agent extracts elements from the knowledge base to give the answer. Conversely, if the agent does not receive a feasible SPARQL, A-Agent provides a direct response with LLM's internal knowledge, following the prompt based on the type of the answer. Additionally, A-Agent will send a retry signal to previous phases if no result is generated. The subtask can be formulated as below:
\begin{equation}
\begin{aligned}
f(S_{ca}) =& Agent_a(LLM, Mem_{ca}, F_{ca}, \\
&Pmt_{ca}, \theta_{ca}, \mathcal{T}), \\ 
Mem_{ca} =& \left[KB\right], \theta_{ca}=\left[Q, f(S_{qs}), f(S_{cls})\right]
\label{eq:eq8}
\end{aligned}
\end{equation}, where $Agent_a$ is the agent as an advisor to perform a comprehensive answering for the question $Q$ with a memory $Mem_{ca}$ of knowledge base, $Pmt_{ca}$ is the prompt for LLM to perform a direct response according to the type of the answer, $f(S_{qs})$ is the final query and $f(S_{cls})$ is the answer type, $\mathcal{T}$ is the maximum times to retry for previous phases if no result is returned from $KB$.

\section{Performance Evaluation}

\subsection{Experimental Settings}
\paragraph{Indexed Knowledge Bases:} The efficacy of our framework is assessed through the collection of two real knowledge bases, specifically DBpedia and YAGO. DBpedia \cite{DBpedia} serves as an accessible knowledge base extracted from Wikipedia, while YAGO \cite{YAGO} is a large knowledge base that includes individuals, cities, nations, and organizations. We index the triples and the mentions of entities and relations in a Virtuoso endpoint and an Elasticsearch server, respectively.

\paragraph{KBQA Benchmark Datasets:} We evaluate our framework on datasets including YAGO-QA, LC-QuAD 1.0, and QALD-9, which have various difficulties in interpreting the questions. These datasets contain questions in English, paired with their respective SPARQL queries, and accurate responses derived from a specific knowledge base. QALD-9 \cite{QALD9} and LC-QuAD 1.0 \cite{LCQUAD} are frequently used to evaluate QA systems with DBpedia. The recently published YAGO-QA in \cite{KGQAN}, features questions accompanied by annotated SPARQL queries sourced from YAGO. The statistics for three benchmarks, along with their associated knowledge bases, are depicted in Table \ref{benchmark}.

\begin{table*}[htbp]
\centering
\begin{tabular}{l|ccccc}
\bottomrule
\multirow{2}{*}{\textbf{Benchmarks}}  & \multicolumn{5}{c}{\textbf{Benchmark Statistics}} 
\\  & 
\multicolumn{1}{c}{\textbf{\#Questions}} & \multicolumn{1}{c}{\textbf{KB}} & \multicolumn{1}{c}{\textbf{\#Triples}} & \multicolumn{1}{c}{\textbf{Virtuoso Size}} & \textbf{ ES size}
\\ \hline      
\multirow{1}{*}{LC-QuAD 1.0} & \multicolumn{1}{c}{1000} & \multicolumn{1}{c}{DBpedia-04} & \multicolumn{1}{c}{397M} & \multicolumn{1}{c}{35.40G} & \multicolumn{1}{c}{1.56G}
\\
\multirow{1}{*}{QALD-9} & \multicolumn{1}{c}{150} & \multicolumn{1}{c}{DBpedia-10} & \multicolumn{1}{c}{374M} & \multicolumn{1}{c}{36.89G} & \multicolumn{1}{c}{1.57G}
\\
\multirow{1}{*}{YAGO-QA} & \multicolumn{1}{c}{100} & \multicolumn{1}{c}{YAGO-4} & \multicolumn{1}{c}{207M} & \multicolumn{1}{c}{24.85G} & \multicolumn{1}{c}{0.54G} 
\\ \bottomrule
\end{tabular}
\caption{The statistics of KBQA benchmarks, including the number of questions number, the number of triples, the size of index in Virtuoso and Elasticsearch.}
\label{benchmark}
\end{table*}

\begin{table*}[htbp]
\centering
\begin{tabular}{lllllllllll}
\bottomrule
\multirow{2}{*}{\textbf{Type}}  & \multirow{2}{*}{\textbf{Frameworks}}  & \multicolumn{3}{c}{\textbf{LC-QuAD 1.0}} & \multicolumn{3}{c}{\textbf{QALD-9}} & \multicolumn{3}{c}{\textbf{YAGO-QA}}
\\  & 
\multicolumn{1}{l}{} &
\multicolumn{1}{l}{\textbf{P}} & \multicolumn{1}{l}{\textbf{R}} & \multicolumn{1}{l}{\textbf{F1}} & \multicolumn{1}{l}{\textbf{P}} & \multicolumn{1}{l}{\textbf{R}} & \multicolumn{1}{l}{\textbf{F1}} & \multicolumn{1}{l}{\textbf{P}} & \multicolumn{1}{l}{\textbf{R}} & \multicolumn{1}{l}{\textbf{F1}}
\\ \hline 
{full-shot} & 
\multirow{1}{*}{gAnswer} & \multicolumn{1}{l}{-} & \multicolumn{1}{l}{-} & \multicolumn{1}{l}{-} & \multicolumn{1}{l}{0.293} & \multicolumn{1}{l}{0.327} & \multicolumn{1}{l}{0.298} & \multicolumn{1}{l}{0.585} & \multicolumn{1}{l}{0.341} & \multicolumn{1}{l}{0.430}
\\
{} & 
\multirow{1}{*}{EDGQA} & \multicolumn{1}{l}{0.505} & \multicolumn{1}{l}{\underline{0.560}} & \multicolumn{1}{l}{\underline{0.531}} & \multicolumn{1}{l}{0.313} & \multicolumn{1}{l}{\underline{0.403}} & \multicolumn{1}{l}{0.320} & \multicolumn{1}{l}{0.419} & \multicolumn{1}{l}{0.408} & \multicolumn{1}{l}{0.414}
\\
{} & 
\multirow{1}{*}{KGQAN} & \multicolumn{1}{l}{\textbf{0.587}} & \multicolumn{1}{l}{0.461} & \multicolumn{1}{l}{0.516} & \multicolumn{1}{l}{\textbf{0.511}} & \multicolumn{1}{l}{0.387} & \multicolumn{1}{l}{\textbf{0.441}} & \multicolumn{1}{l}{0.485} & \multicolumn{1}{l}{\underline{0.652}} & \multicolumn{1}{l}{0.556}
\\  \hline
{few-shot} & 
\multirow{1}{*}{GPT-3.5} & \multicolumn{1}{l}{0.269} & \multicolumn{1}{l}{0.251} & \multicolumn{1}{l}{0.266} & \multicolumn{1}{l}{0.240} & \multicolumn{1}{l}{0.217} & \multicolumn{1}{l}{0.228} & \multicolumn{1}{l}{0.171} & \multicolumn{1}{l}{0.142} & \multicolumn{1}{l}{0.155}
\\
{} & 
\multirow{1}{*}{GPT-4} & \multicolumn{1}{l}{0.336} & \multicolumn{1}{l}{0.344} & \multicolumn{1}{l}{0.340} & \multicolumn{1}{l}{0.250} & \multicolumn{1}{l}{0.249} & \multicolumn{1}{l}{0.249} & \multicolumn{1}{l}{0.193} & \multicolumn{1}{l}{0.190} & \multicolumn{1}{l}{0.191}
\\  
{} & 
\multirow{1}{*}{Triad-GPT3.5} & \multicolumn{1}{l}{0.490} & \multicolumn{1}{l}{0.519} & \multicolumn{1}{l}{0.504} & \multicolumn{1}{l}{0.293} & \multicolumn{1}{l}{0.302} & \multicolumn{1}{l}{0.297} & \multicolumn{1}{l}{\underline{0.660}} & \multicolumn{1}{l}{0.639} & \multicolumn{1}{l}{\underline{0.649}}
\\
{} & 
\multirow{1}{*}{\textbf{Triad-GPT4}} & \multicolumn{1}{l}{\underline{0.561}} & \multicolumn{1}{l}{\textbf{0.568}} & \multicolumn{1}{l}{\textbf{0.564}} & \multicolumn{1}{l}{\underline{0.408}} & \multicolumn{1}{l}{\textbf{0.425}} & \multicolumn{1}{l}{\underline{0.416}} & \multicolumn{1}{l}{\textbf{0.690}} & \multicolumn{1}{l}{\textbf{0.664}} & \multicolumn{1}{l}{\textbf{0.677}}
\\ \bottomrule
\end{tabular}
\caption{The performance of our proposed Triad on three benchmarks, comparing with traditional KBQA systems (full-shot) and pure LLM (few-shot) baselines. The optimal and suboptimal scores are highlighted with bold and underlined text, respectively.}
\label{performance}
\end{table*}

\paragraph{Baseline Methods:} We evaluate Triad against traditional KBQA systems such as KGQAN \cite{KGQAN}, EDGQA \cite{EDGQA} and gAnswer \cite{gAnswer}. This comparison shows how our LLM-based agent framework can rival full-shot systems with just a few examples. Additionally, we contrast our framework with pure GPT models like GPT-3.5 Turbo and GPT-4 \footnote{https://platform.openai.com/docs/models} to exhibit Triad's architectural performance. We treat these foundation models as few-shot methods to answer the questions referring to some examples.

\paragraph{Implementation Details:} Triad is implemented with Python 3.9. We incorporate LLM capabilities to our multi-role agent via OpenAI's API services. The names of entities and relations from knowledge bases are indexed in an ElasticSearch 7.5.2 server for text matching. All triples are imported into an SPARQL endpoint of Virtuoso 07.20.3237 for retrieval. Triad requires four hyperparameters: the number of examples G-Agent uses for subtask learning, the number of candidates D-Agent selects for entity and relation linking, and the retry times for handling non-response SPARQLs. The optimal values for these parameters are 3, 2, 2, and 3, respectively. The framework and its variants are tested five times on each benchmark, with the average scores reported as the final results. For traditional systems, we report the results recorded in their papers. For pure LLM baselines, we write prompts to hire an LLM to answer questions directly referring to examples, and then link the mentions from the responses to the URIs in our indexed knowledge bases via built-in similarity search.

\subsection{Performance Comparison}
The performance of \textbf{Triad} compared to traditional KBQA systems and pure LLM generation methods is shown in Table \ref{performance}. Evaluation metrics precision(P), recall(R), and F1-score(F1) are reported. We can observe from the experimental results that:

\paragraph{Few-shot can be competitive with full-shot.} Our multi-role LLM-based agent framework, though executing a few-shot prompt learning, exhibits competitive performance with cutting-edge full-shot KBQA systems.
\paragraph{Underlying capability matters.} The use of GPT-4 as the core in an LLM-based agent significantly outperforms GPT-3.5 on all benchmarks, demonstrating the importance of the underlying capabilities of an agent.
\paragraph{Explicit knowledge is necessary.} Pure LLM models with GPT-3.5 and GPT-4 display deficiencies in generating accurate responses without an auxiliary knowledge base as a memory for intermediary steps such as URI linking.
\paragraph{Performance varies with complexity.} Triad demonstrates superior results on the LC-QuAD and YAGO-QA benchmarks compared to QALD-9, due to an increasing failure in response to complex questions, which will be discussed later.

\subsection{Study on Capabilities of Agent Roles }
We assess the efficacy of G-Agent with various other language models as the core. The framework without \textbf{G-task} uses the text-davinci-002, which is not as powerful as GPT-3.5 and GPT-4 in solving many tasks, and the one without \textbf{G-chat} uses text-davinci-003 to eliminate the chat and alignment abilities. We test the ability of D-Agent without \textbf{D-uri} and \textbf{D-query} by replacing the URI selection and query selection with URI matching and query generation, respectively. We evaluate the contribution of A-Agent eliminating \textbf{A-llm} and \textbf{A-fact} by responding to questions without using LLM's assistance or use an LLM to answer Boolean questions for auxiliary rather than single-fact questions. The F1 results of the role ablation experiments on two representative datasets are shown in Table \ref{ablation}. The results indicate that every component pertaining to each role contributes to the overall performance. More specifically, a G-Agent that employs a less powerful LLM as its core can drastically undermine performance. D-Agent assumes a more pivotal role during the linking phase compared to the query construction phase. A-Agent, on the other hand, proves to be an efficient solution for managing situations where SPARQL results are absent.

\begin{table}[htbp]
\centering
\begin{tabular}{cccc}
\bottomrule
\textbf{G-task} & \textbf{G-chat} & \textbf{LC-QuAD 1.0} & \textbf{ QALD-9 }\\
\hline
\XSolidBrush & \XSolidBrush & 0.343 & 0.159 \\
\Checkmark & \XSolidBrush & 0.443 & 0.248 \\
\Checkmark & \Checkmark & 0.564 & 0.416 \\
\bottomrule
\textbf{D-uri} & \textbf{D-query} & \textbf{LC-QuAD 1.0} & \textbf{ QALD-9 }\\
\hline
\XSolidBrush & \Checkmark & 0.274 & 0.210 \\
\Checkmark & \XSolidBrush & 0.431 & 0.301 \\
\Checkmark & \Checkmark & 0.564 & 0.416 \\
\bottomrule
\textbf{A-llm} & \textbf{A-fact} & \textbf{LC-QuAD 1.0} & \textbf{ QALD-9 }\\
\hline
\XSolidBrush & \XSolidBrush & 0.459 & 0.382 \\
\Checkmark & \XSolidBrush & 0.473 & 0.385 \\
\Checkmark & \Checkmark & 0.564 & 0.416 \\
\bottomrule
\end{tabular}
\caption{ Study on the roles of LLM-based agent by eliminating an element or downgrading the capability.}
\label{ablation}
\end{table}

\subsection{Analysis of Role Hyperparameters}
We concentrate on three hyperparameters of roles, including the number of examples ($\mathcal{N} \in \{1,2,3\}$) provided for G-Agent to learn subtasks, the number of URI candidates ($\mathcal{K} \in \{(1,1),(1,2),(2,2),(2,3)\}$) selected by D-Agent for query construction, and the number of retry times ($\mathcal{T} \in \{1,2,3\}$) launched by A-Agent when there is no response. Table \ref{hyper} presents the F1 results of Triad's performance, employing three hyperparameters on two benchmarks. We discover that:

\begin{table}[htbp]
\centering
\begin{tabular}{lcc}
\bottomrule
\textbf{Triad Variants} & \textbf{LC-QuAD 1.0} & \textbf{ QALD-9 }\\
\hline
Triad-1-Shot & 0.556 & 0.376 \\
Triad-2-Shot & 0.511 & 0.402 \\
Triad-3-Shot & 0.564 & 0.416 \\
\hline
Triad-Top1-1 & 0.528 & 0.281 \\
Triad-Top1-2 & 0.562 & 0.375 \\
Triad-Top2-2 & 0.564 & 0.416 \\
Triad-Top2-3 & 0.558 & 0.384 \\
\hline
Triad-1-Try & 0.529 & 0.375 \\
Triad-2-Tries & 0.561 & 0.407 \\
Triad-3-Tries & 0.564 & 0.416 \\
\bottomrule
\end{tabular}
\caption{ Performance evaluation on three hyperparameters that related to each role of an LLM-based agent.}
\label{hyper}
\end{table}

\paragraph{Quality is more important than quantity.} More examples provided to G-Agent do not always improve the performance. The efficacy of G-Agent is significantly influenced by the quality of examples.
\paragraph{More options may harm the result.} Choosing more candidate URIs for entities and relations could potentially disrupt subsequent query phases, thus affecting overall performance.
\paragraph{More chances benefits the framework.} Persistently attempting to construct and execute SPARQL queries is an effective strategy that improves the probability of obtaining accurate answers. Considering the efficiency of overall execution, we set the maximum retry times as 3 in practice. 

\subsection{Analysis of Linking Recall}
The process of linking is a relatively complex subtask in both the Text2SQL and the KBQA process \cite{BIRD}. Calculating the recall ratio of accurate URIs using D-Agent provides clarity on which step most adversely impacts performance. In the entity linking phase, considering all URIs of entities in the testing set as the ground truth of the linking results, 80. 75\% of the correct URIs are contained from the output of the entity matching filter in D-Agent and 70. 50\% of the correct URIs are retained from the entity selection performed by the LLM in D-Agent. Whereas, in the relation linking phase, only 52. 54\% of the correct relation URIs survive from the selection of LLM, which indicates a greater difficulty in relation linking.

\subsection{Study on Complex Cases}
Despite the impressive performance of Triad in certain benchmarks, notable deficiencies remain in its ability to understand questions and generate queries for complex questions. A critical analysis of unsuccessful cases in QALD-9, which has the lowest F1 score, has revealed three primary reasons for this failure, as detailed below:

\begin{table}[htbp]
\centering
\begin{tabular}{lcl}
\bottomrule
\makecell[c]{\textbf{Fail Reason}} & \makecell[c]{\textbf{Ratio}} & \makecell[c]{\textbf{Example}}\\
\hline
\makecell[c]{Complex \\ Syntax} & 20\% & \makecell[c]{Q42: Which \\ countries have \\ places with more \\ than two caves?} \\ \hline
\makecell[c]{Unexploited \\ Semantics} & 17\% & \makecell[c]{Q199: Give me all \\ Argentine films.} 
\\ \hline
\makecell[c]{Implicit \\ Reasoning} & 5\% & \makecell[c]{Q133: What are \\ the names of the \\ Teenage Mutant \\ Ninja Turtles?} \\
\bottomrule
\end{tabular}
\caption{ The major reasons of complexity that result in failures, with their corresponding ratios of occurrence in failed cases.}
\label{complex}
\end{table}

\paragraph{Complex Syntax} signifies that advanced SPARQL queries incorporate keywords such as \textit{GROUP BY} and \textit{HAVING}. These terms augment the error propensity in the generation of SPARQL templates such as the example: \textit{Which frequent flyer program has the most airlines?}
\paragraph{Unexploited Semantics} indicates that semantics of an implicit entity should be comprehended in order to exclude irrelevant URIs. In the example \textit{Give me all Argentine films}, the meaning of \textit{films} should be used to narrow down the scope of potential entities in order to eliminate unrelated answers.
\paragraph{Implicit Reasoning} presents a challenge that requires a deeper level of traversal by the framework to deduce accurate results from the posed question. For example, another failure question, \textit{How many grand-children did Jacques Cousteau have?}, the term \textit{grand-children} must be interpreted to \textit{son of son} to ensure an accurate response.

\subsection{Cost Comparison and Analysis}
According to our evaluation on the three datasets, the average cost of running a single case is 0.007 USD on average using Triad-GPT3.5 and 0.05 USD on average using Triad-GPT4. Specifically, most API calls occur in the phases of URL linking and comprehensive answering. Meanwhile, traditional KBQA baselines require a lot of training data and local training resources to achieve the SOTA performance, whereas Triad follows a zero- or few-shot manner to save computational cost locally. Furthermore, as shown in Section 4.4, in practice, adjusting the hyperparameters can make the cost as low as possible while preserving overall performance. As the cost of LLM services decreases, the value of Triad will increase accordingly.

\section{Related Work}

\subsection{SPARQL-based and LLM-based KBQA}
Traditional KBQA methods transform natural language queries into SPARQL requests for data extraction. Specific models are employed either for question understanding or URI linking, utilizing domain-based training datasets. \citet{gAnswer} introduces a semantic query graph to structurally represent the natural language query, thereby simplifying the task into a subgraph matching problem. \citet{EDGQA} proposes an entity description graph to represent natural language queries for question parsing and element linking. \citet{KGQAN} restructures the question parsing task as a text generation issue using a sequence-to-sequence model. With the advent of LLMs, certain phases of KBQA can be enhanced with LLM-integrated methods. \citet{PromptING} aims to augment LLM-based QA tasks with pertinent facts extracted from knowledge bases, offering a fully zero-shot architecture. \citet{McL-KBQA} leverages the general applicability of LLMs to filter linking candidates by making selections via few-shot in-context learning. \citet{ChatGPT-KBQA} provides a thorough comparison between LLMs and QA systems, recommending further studies to improve KBQA with LLM capabilities. However, apart from the above studies, our study proposes a complete framework incorporating both an LLM and few-shot learning across all KBQA phases from a systematic perspective.

\subsection{LLM-based Agents for Complex Tasks}
LLMs have recently gained significant attention due to their ability to approximate human-level intelligence. This has led to numerous studies focusing on LLM-based agents. A recent survey\cite{AgentSurvey} proposes a unified architecture for LLM-based agents, which consists of four modules that include profile, memory, plan, and action. CHATDB\cite{CHATDB} employs an LLM controller to generate SQL instructions, which allows for symbolic memory and complex multi-hop reasoning. ART\cite{ART} uses a frozen LLM to generate reasoning steps and further integrates tools for new tasks with minimal human intervention. Toolformer\cite{Toolformer} takes a different approach by training an LLM to plan and execute tools for the next token prediction by learning API calls generation. ReAct\cite{ReAct} focuses on overcoming LLM hallucination by interacting with external knowledge bases, thus generating interpretable task-solving strategies. CodeAgent\cite{tang2024collaborative} designs a multi-agent collaboration system across four phases in a code review process. Divergent from the aforementioned studies, our framework concentrates on the solving KBQA tasks by introducing a multi-role LLM-based agent that specializes in various subtasks distributed across different phases.

\section{Conclusion}
In this study, we aim to bridge the gap between KBQA tasks and the investigation of LLM-based agents. We introduce \textbf{Triad}, a framework to address the KBQA task through an LLM-based agent acting as multiple roles, including a generalist capable of mastering diverse tasks given minimal examples, a decision-maker concentrating on option analysis and candidate selection, and an advisor skilled in answering questions with the aid of both external and internal knowledge. Triad achieves the best or competitive performance across three benchmark datasets compared to traditional KBQA systems and pure LLM models. In future research, we plan to broaden our framework to handle more intricate questions, such as multi-hop reasoning, and exploring the integration between our framework and retrieval-augmented generation.

\section*{Limitations}
The limitation of our research lies in following aspects: (1) In terms of data, a broader range of QA datasets needs to be evaluated, encompassing datasets from different domains, languages, and difficulty levels. (2) In terms of model, more LLMs need to be evaluated, including open-source and commercial models from different organizations and on various scales. (3) In terms of framework, more types of agent collaboration methods can be explored to solve KBQA problems. 

\section*{Ethics Considerations}
All datasets utilized in this study are publicly available and we have adhered to ethical considerations by not introducing additional information as input during LLM text generation.

\section*{Acknowledgements}
This work is supported by the "Pioneer" and "Leading Goose" R\&D Program of Zhejiang (Grant No. 2023C01152).

\bibliography{anthology,custom}

\clearpage
\appendix
\section{Response Time Analysis}
\begin{figure}[htbp]
    \begin{center}
    \includegraphics[width=.49\textwidth]{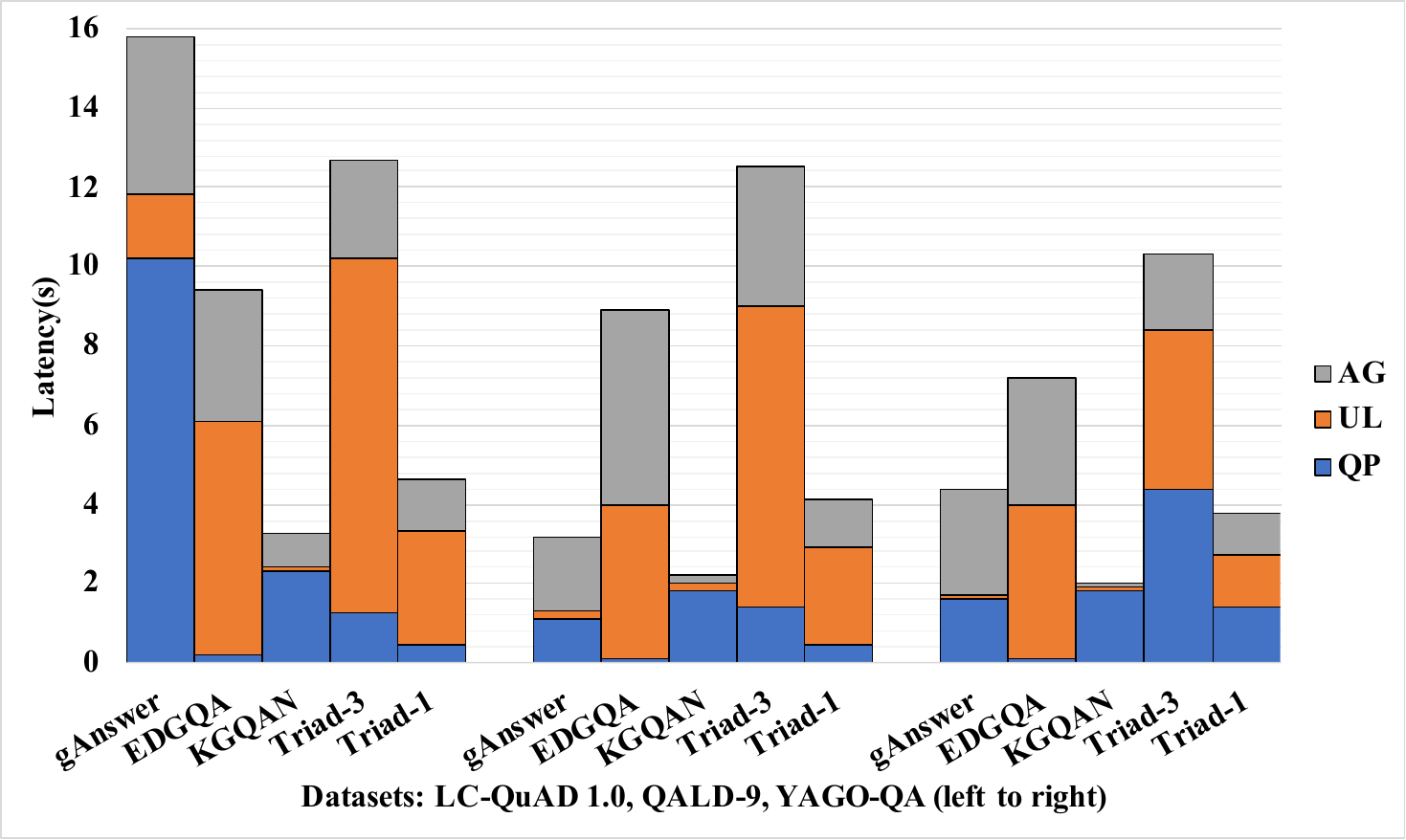} 
    \caption{Response time of traditional KBQA systems and Triad on three datasets. Each bar shows average response time of a particular phase of KBQA.}
    \label{efficiency}
    \end{center}
\end{figure}

We analyze various QA frameworks in response time to a question. The average latency of each phase including question parsing (QP), URI linking (UL), and answer generation (AG) for each knowledge base is reported. We randomly select 10 samples from each dataset for evaluation to obtain the average response times for Triad-1 and Triad-3, which represent retrying three times and generating an answer in one go, respectively, during the answer generation phase. The comparison between traditional QA systems and Triad is shown in Figure \ref{efficiency}. Triad generally shows a competitive time-consuming performance to latest traditional QA systems. Specifically, compared to other phases, URL linking consumes more time due to the need to invoke LLM multiple times. Moreover, according to Section 4.4, with smaller retry times of A-Agent, Triad can significantly reduce time cost while only causing slight performance degradation, revealing the advantages of our framework in balancing performance and efficiency.

\section{Role Performance on YAGO-QA}
We choose LC-QuAD 1.0 and QALD-9 as our two representative datasets in Section 4.3, as the questions among them vary in difficulty, and the tasks in these two datasets are relatively more challenging than YAGO-QA. We provide the performance of agent roles on YAGO-QA in Table \ref{ablation2}, which shows a consistent result with other datasets in Table \ref{ablation}.

\begin{table}[htbp]
\centering
\begin{tabular}{ccc}
\bottomrule
\textbf{G-task} & \textbf{G-chat} & \textbf{YAGO-QA} \\
\hline
\XSolidBrush & \XSolidBrush & 0.427 \\
\Checkmark & \XSolidBrush & 0.553 \\
\Checkmark & \Checkmark & 0.677 \\
\bottomrule
\textbf{D-uri} & \textbf{D-query} & \textbf{YAGO-QA}\\
\hline
\XSolidBrush & \Checkmark & 0.346 \\
\Checkmark & \XSolidBrush & 0.534 \\
\Checkmark & \Checkmark & 0.677 \\
\bottomrule
\textbf{A-llm} & \textbf{A-fact} & \textbf{YAGO-QA}\\
\hline
\XSolidBrush & \XSolidBrush & 0.626 \\
\Checkmark & \XSolidBrush & 0.647 \\
\Checkmark & \Checkmark & 0.677 \\
\bottomrule
\end{tabular}
\caption{Performance of roles of LLM-based agent by eliminating an element or downgrading the capability.}
\label{ablation2}
\end{table}

\section{Prompts Provided to LLMs of G-Agent for Solving Various Subtasks in KBQA}

The prompt given to LLMs of $Agent_g$ to perform triplet extraction from the question $Q$ is as follows:
\begin{center}
\fcolorbox{black}{gray!10}{\parbox{1.0\linewidth}{You are an assistant to \textit{\textbf{identify triples}} within a provided sentence. Please adhere to the following \textbf{guidelines:} \\

1. Triples should be structured in the format <entity1, relation, entity2>. \\
2. The sentence must contain at least one triple, so you should provide at least one. \\
3. Entities should represent the smallest semantic units and should not contain descriptive details. \\
4. Entities can take the form of explicit or implicit references. Explicit entities refer to specific named resources, whereas implicit entities are less certain. \\
5. When an entity is implicit, utilize a variable format such as '?variable' to denote it, for example, '?location' or '?person'. \\

\textbf{Here are some examples:} \\
Which city's founder is John Forbes? : <?city, foundeer, John Forbes>\\
How many races have the horses bred by Jacques Van't Hart participated in? : <?horse, participated in, ?race> <?horse, breeder, Jacques Van't Hart> \\
Is camel of the chordate phylum? : <camel, phylum, chordate> \\

\textbf{Sentence:} <Question Sentence> \\
\textbf{Output:} }}
\end{center}

The prompt given to LLMs of $Agent_g$ for SPARQL template generation is as follows:
\begin{center}
\fcolorbox{black}{gray!10}{\parbox{1.0\linewidth}{
You are an assistant to \textit{\textbf{generate a SPARQL query}} to address a specific question. Here are the \textbf{guidelines to follow:} \\

1. Ensure that the resulting SPARQL query is designed to answer the provided question. \\
2. Adhere to the commonly accepted SPARQL standards when generating the query. \\
3. Make an effort to leverage the information provided to assist in the creation of the SPARQL query. \\
4. Strive to keep the generated SPARQL query as straightforward as possible. \\
5. Avoid including 'PREFIX' or ':' in the SPARQL query. \\
6. Enclose condition entities and predicates within angle brackets, such as <entity> or <predicate>. \\
7. Maintain the original order of the given triples without altering their sequence. \\

\textbf{Question:} <question sentence>\\
\textbf{Triplets:} <extracted triplets>\\
\textbf{Output:} }}
\end{center}

The prompt given to LLMs of $Agent_g$ for question type classification is as follows:
\begin{center}
\fcolorbox{black}{gray!10}{\parbox{1.0\linewidth}{You are an assistant to \textit{\textbf{determine the specific type}} of a given question according to the following guidelines: \\

1. You must determine the most probable question type for the input question. \\
2. The type of question should be enclosed within angle brackets, denoted as '<' and '>'. \\
3. Possible question types include: <count>, <select>, and <yes or no>. \\

\textbf{Question:} <question sentence> \\
\textbf{Output:} }}
\end{center}

\section{Prompts Provided to LLMs of D-Agent for Solving Selection Subtasks in KBQA}

The prompt given to LLMs of $Agent_d$ for candidate entities selection is as follows:
\begin{center}
\fcolorbox{black}{gray!10}{\parbox{1.0\linewidth}{You are an assistant to \textit{\textbf{select <K> URIs}} from a provided list of possible URIs for a specified entity, following these \textbf{guidelines:} \\

1. Identify the <K> most appropriate URIs from the given list that best represent the entity in question. \\
2. Seek to understand the semantic information associated with the specified entity by examining the provided question. \\
3. The output should consist of <K> URIs chosen from the provided list of possible URIs. \\
4. Simply output these <K> target URIs, each on a separate line, without providing any additional explanations. \\

\textbf{Sentence:} <question sentence>\\
\textbf{Entity:} <entity mention> \\
\textbf{Possible entity URIs:} <Entity URI list> \\
\textbf{Output:} }}
\end{center}

The prompt given to LLMs of $Agent_d$ for candidate relation selection is as follows:
\begin{center}
\fcolorbox{black}{gray!10}{\parbox{1.0\linewidth}{
You are an assistant tasked with \textit{\textbf{selecting the <K> relation URIs}} between entities mentioned in a sentence. Here are the \textbf{guidelines:} \\

1. The two entities are listed one after the other, without a specific order. \\
2. Use the provided sentence to discern the semantic meaning of these entities. \\
3. The potential relation URIs are listed one by one. \\
4. Your output should consist of a maximum of <K> possible relation URIs, although you may also output fewer if appropriate. \\
5. Ensure that your output is organized, prioritizing the most likely relationship first. \\
6. Provide a list of no more than <K> relation URIs (each on a separate line if there are multiple) without any additional descriptions. \\

\textbf{Sentence:} <question sentence>\\
\textbf{Entities:} <entity pair>\\
\textbf{Possible relation URIs:} <URI list>\\
\textbf{Output:} }}
\end{center}

The prompt given to LLMs of $Agent_d$ for the final SPARQL selection is as follows:
\begin{center}
\fcolorbox{black}{gray!10}{\parbox{1.0\linewidth}{
You are an assistant to \textit{\textbf{select an appropriate SPARQL query}} from the provided list in order to respond to a specific question. Please adhere to the following \textbf{guidelines:} \\

1. Select the most suitable SPARQL query from the given query list to address the question. \\
2. Select a SPARQL query solely from the provided list; avoid crafting your own SPARQL query. \\
3. The selected SPARQL query must be applicable to answer the given question. \\

\textbf{Sentence:} <question sentence>\\
\textbf{SPARQL candidates:} <SPARQLs to choose>\\
\textbf{Output:} }}
\end{center}

\section{Prompts Provided to LLMs of A-Agent for Solving Answering Subtask in KBQA}

The prompt given to LLMs of $Agent_a$ to generate a yes or no answer for the give question is as follows:
\begin{center}
\fcolorbox{black}{gray!10}{\parbox{1.0\linewidth}{
You are an assistant to \textit{\textbf{answer a yes-or-no question}}. Please adhere to the following \textbf{guidelines:} \\

1. If you believe that the answer is yes, provide an output of 'True'. If not, provide an output of 'False'.\\
2. Please do not include additional information or explanations in your response.\\

\textbf{Sentence:} <question sentence>\\
\textbf{Output:} }}
\end{center}

The prompt given to LLMs of $Agent_a$ to generate a single-fact answer for the give question is as follows:
\begin{center}
\fcolorbox{black}{gray!10}{\parbox{1.0\linewidth}{
You are an assistant to \textit{\textbf{answer a question}}. Please adhere to the following \textbf{guidelines:} \\

1. The answer to the question is a single entity. \\
2. You should just output the full expression of the answer without any punctuation. \\
3. Do not output any other description. \\

\textbf{Sentence:} <question sentence>\\
\textbf{Output:} }}
\end{center}

%\label{sec:appendix}
%This is an appendix.

\end{document}